%% file: ALTMAS_RobustML_ICLRSubmission.tex
\documentclass{article} 
\usepackage{RobustML_iclr2021_conference,times}

\input{math_commands.tex}

\usepackage{hyperref}
\usepackage{url}
\usepackage{amsmath}
\usepackage{amssymb}
\usepackage{amsthm}
\usepackage{amsfonts}
\usepackage{mathrsfs}
\usepackage[normalem]{ulem}
\usepackage{algorithm}
\usepackage{algorithmic}
\usepackage{enumitem}
\usepackage[xindy,acronym,toc,smallcaps,nomain]{glossaries}

\usepackage{color}
\usepackage{graphicx}
\usepackage{xcolor}
\newcommand*\diff{\mathop{}\!\mathrm{d}}

\newtheorem{remark}{Remark}[section]

\makeglossaries

\newacronym{ALT-MAS}{altmas}{placeholder}

\title{ALT-MAS: A Data-Efficient Framework for Active Testing of Machine Learning Algorithms}

\author{Huong Ha\thanks{Correspondence to: \texttt{huong.ha@rmit.edu.au}} \\
RMIT University\\
Australia \\
\And
Sunil Gupta \\
Deakin University \\
Australia
\And
Santu Rana \\
Deakin University \\
Australia
\And
Svetha Venkatesh \\
Deakin University \\
Australia
}

\iclrfinalcopy 
\begin{document}

\setlength{\belowdisplayskip}{3pt}

\maketitle

\begin{abstract}
Machine learning models are being used extensively in many important areas, but there is no guarantee a model will always perform well or as its developers intended. Understanding the correctness of a model is crucial to prevent potential failures that may have significant detrimental impact in critical application areas. In this paper, we propose a novel framework to efficiently test a machine learning model using only a small amount of labeled test data. The idea is to estimate the metrics of interest for a model-under-test using Bayesian neural network (BNN). We develop a novel data augmentation method helping to train the BNN to achieve high accuracy. We also devise a theoretic information based sampling strategy to sample data points so as to achieve accurate estimations for the metrics of interest. Finally, we conduct an extensive set of experiments to test various machine learning models for different types of metrics. Our experiments show that the metrics estimations by our method are significantly better than existing baselines.
\end{abstract}

\section{Introduction}
\label{introduction}

Today, supervised machine learning models are employed across sectors to assist humans in making important decisions. Understanding the correctness of a model is thus crucial to avoid potential (and severe) failures. In practice, however, it is not always possible to accurately evaluate the model's correctness using the held-out training data in the development process \citep{Sawade2010}. Consider a hospital that buys an automated medical image classification system. The supplier will provide a performance assessment, but this evaluation may not hold in this new setting as the supplier and the hospital data distributions may differ. Similarly, an enterprise that develops a business prediction system might find that the performance changes significantly over time as the input distribution shifts from the original training data. In these cases, the model performance needs to be re-evaluated as the assessments provided from the supplier or from the development process can be inaccurate. To accurately evaluate the model performance, new labeled data points from the deployment area are needed. But the process of labelling is expensive as one would usually need a large number of test instances. \textit{Thus the open question is how to test the performance of a machine learning model (model-under-test) with parsimonious use of labeled data from the deployment area.} 

This work focuses on addressing this challenge treating the model-under-test as a black-box as in common practice one only has access to the model outputs. We propose a \textit{data-efficient framework} that can assess the performance of a black-box model-under-test on any metric (that is applicable for black-box models) without prior knowledge from users. Furthermore, our framework can estimate multiple metrics simultaneously. The motivation for evaluating one or multiple metrics is inspired by the current practice of users who need to assess the model-under-test on one or varied aspects that are important to them. For instance, for a classification system, the user might want to solely check the overall accuracy or simultaneously check the overall accuracy, macro-precision (recall) and/or the accuracies of some classes of interest.

To achieve sample efficiency, we formulate our testing framework as an active learning (AL) problem \citep{Cohn1996}. First, a small subset of the test dataset is labeled, and a surrogate model is learned from this subset to predict the ground truth of unlabeled data points in the test dataset. Second, an acquisition function is constructed to decide which data point in the test dataset should be chosen for labelling. The data point selected by the acquisition function is sent to an external oracle for labelling, and is added to the labeled set. The process is conducted iteratively until the labelling budget is depleted. The metrics of interest are estimated using the learned surrogate model.

With this framework, one choice is to use a standard AL method to learn a surrogate model that accurately predicts the labels of all data points in the test dataset, however, this choice is not optimal. To efficiently estimate the metrics of interest, the surrogate model should not need to accurately predict the labels of all data points; it only needs to accurately predict the labels of those data points that contribute significantly to the accuracy of the metric estimations. For our active testing framework, we first propose a data augmentation method that helps to train the surrogate model to achieve high accuracy. Second, we derive a theoretic information based acquisition function that can select data points for whom labels should be acquired so as to enable maximal estimation accuracy of the metric of interest. We then generalize our framework to be able to work with multiple metrics. Finally, we demonstrate the efficacy of our proposed testing framework using various models-under-test and a wide range of metric sets. In summary, our main contributions are:
\begin{enumerate}
\item
ALT-MAS, a data-efficient testing framework that can accurately estimate the performance of a machine learning model;
\item
A novel data augmentation strategy to train BNNs to achieve high accuracy, and a novel sampling methodology to estimate the metrics of interest accurately and efficiently; and,
\item
Demonstration of the empirical effectiveness of our proposed machine learning testing framework on various models-under-test for a wide range of metrics.
\end{enumerate}

\paragraph{Related work} There have been various research works tackling the problem of evaluating machine learning model performance using limited labeled data. However, to the best of our knowledge, the methods proposed in all these works, except \citet{2011Druck} and \citet{Welinder2013}, can only work with some specific metrics such as accuracy and precision-recall curve. The approaches in \citet{2011Druck} and \citet{Welinder2013} can estimate any metrics like ours, however, they require information from the classifier internal structure, particularly the softmax outputs. Further details regarding related work are in App. \ref{sec:app-relatedwork}.

\section{Problem Formulation}
\label{problem_form}
\vspace{-0.2cm}
Let us assume we are given a black-box model-under-test $\mathcal{A}$ that gives the prediction $\mathcal{A}(x)$ for an input $x$, with $\mathcal{A}(x) \in \mathcal{C}=\{1, \dots, C\}$. Let us also assume we have access to (i) an unlabeled test dataset $\mathcal{X} = \{x_i\}_{i=1}^N$, and, (ii) an oracle that can provide the label $y_x$ for each input $x$ in $\mathcal{X}$. Given a set of performance metrics $\{Q_k \}_{k=1}^K, Q_k: \mathbb{R}^N \times \mathbb{R}^N \rightarrow \mathbb{R}$, the goal is to \textit{efficiently} estimate the values of these metrics, $\{ Q_k^*\}_{k=1}^K$, when evaluating the model-under-test $\mathcal{A}$ on the test dataset $\mathcal{X}$. That is, we aim to estimate,
\begin{equation} \label{eq-prob}
Q_k^* = Q_k(\mathcal{A}_{\mathcal{X}}, \mathcal{Y}_{\mathcal{X}}),\quad k=1,\dots,K,
\end{equation}
with $\mathcal{A}_{\mathcal{X}} = \{ \mathcal{A}(x) \}_{x \in \mathcal{X}}$ and $\mathcal{Y}_{\mathcal{X}}= \{y_x\}_{x \in \mathcal{X}}$, using the minimal number of oracle queries. Note we focus on classifiers because they are common supervised learning models and also the target models of most machine learning testing papers \citep{2011Druck, Welinder2013, SabharwalS17, Jiang2018, Zhang2019}.

\section{ALT-MAS: Active Testing with Metric-Aware Sampling}

\vspace{-0.2cm}
For our active testing framework, BNN is chosen as the surrogate model due to its effectiveness and scalability. In this section, we propose methods to (i) augment the training data so that the BNN can achieve high accuracy from a limited number of labeled data (\textbf{Section \ref{sec_surmodel}}), (ii) estimate the metrics of interest given the BNN (\textbf{Section \ref{sec_metest}}), and (iii) sample the most informative data point to maximize the estimation accuracy of a specific metric (\textbf{Section \ref{sec:sample-one}}) or a set of metrics (\textbf{Section \ref{sec-samplelist}}). 

\subsection{Data Augmentation Strategy} \label{sec_surmodel}
\vspace{-0.2cm}
At each iteration $t$, we train a binary classifier $\mathcal{C}_{\eta}$ aiming to predict the data points (in the test dataset $\mathcal{X}$) whose model-under-test predictions agree with the ground-truth. Using the predictions by $\mathcal{C}_{\eta}$, we then construct an \textit{augmented labeled set} $\mathcal{S}_l^t = \lbrace \mathcal{X}_S^t, \mathcal{Y}_{\mathcal{X}_S^t} \rbrace$ where $\mathcal{X}_S^t$ are all data points in $\mathcal{X}$ which $\mathcal{C}_{\eta}$ identifies that the model-under-test predictions are accurate, and $\mathcal{Y}_{\mathcal{X}_S^t}$ are the corresponding model-under-test outputs of $\mathcal{X}_S^t$. The BNN is then trained using both the labeled set $\mathcal{D}_{l}^t$ and the augmented labeled set $\mathcal{S}_l^t$. Further details on how to train the binary classifier is in App. \ref{sec:app-data-aug}.

\subsection{Metric Estimation with Bayesian Neural Network} \label{sec_metest}
\vspace{-0.2cm}
At iteration $t$, using the predicted posterior distribution $q_{\theta}(\omega \vert \mathcal{D}_{l}^t, \mathcal{S}_l^t)$ provided by the trained BNN with weights $\omega$, we can approximate the value of a metric $Q_k$ as,
\begin{equation} \label{eq:met-est}
\begin{aligned}
\hat{Q}_{k}^t & = \int Q_k(\mathcal{A}_{\mathcal{X}}, [\mathcal{Y}_{\mathcal{X}_{l}^t}, \hat{\mathcal{Y}}_{\mathcal{X}_{ul}^t,\omega}]) q_{\theta}(\omega \vert \mathcal{D}_{l}^t, \mathcal{S}_l^t) \diff \omega \approx \dfrac{1}{M} \sum\nolimits_{j=1}^M Q_k(\mathcal{A}_{\mathcal{X}}, [\mathcal{Y}_{\mathcal{X}_{l}^t}, \hat{\mathcal{Y}}_{\mathcal{X}_{ul}^t,\hat{\omega}_j}]),
\end{aligned}
\end{equation}
where $\mathcal{X}_{l}^t, \mathcal{X}_{ul}^t$ denote the sets of labeled and unlabeled data points in $\mathcal{X}$ respectively, $\hat{\mathcal{Y}}_{\mathcal{X}_{ul}^t, \omega}$ denotes the predicted labels of ${\mathcal{X}_{ul}^t}$ using the BNN weights $\omega$, and $\{\hat{\omega}_j\}_{j=1}^M$ are $M$ stochastic forward passes from the BNN predicted posterior distribution $q_{\theta}(\omega \vert \mathcal{D}_{l}^t, \mathcal{S}_l^t)$.

\subsection{Sampling Methodology for a Single Metric} \label{sec:sample-one}
\vspace{-0.2cm}
At iteration $t$, the sampling process aims to select a data point $x_t^*$ to label so as to maximally increase the metric estimation accuracy. This can be considered equivalent to sampling the data point $x_t^*$ such that knowing its label results in maximal uncertainty reduction for the metric estimation. To identify $x_t^*$, our idea is to (i) evaluate how much the uncertainty of each data point's label contributes to the uncertainty of the metric estimation, and, (ii) sample the data point causing the highest uncertainty in the metric estimation. \textbf{To solve (i)}, we define a new random variable (r.v.) $\tilde{Q}_k (x)$ as follows,
\begin{equation} \label{eq-Qkx}
\begin{aligned}
\tilde{Q}_k (x) &= \mathbb{E}_{\hat{\mathcal{Y}}_{\mathcal{X}_{ul}^t \setminus x} \sim p(\hat{\mathcal{Y}}_{\mathcal{X}_{ul}^t \setminus x} \vert \mathcal{X}_{ul}^t \setminus x, \mathcal{D}_l^t)} \ \lbrack Q_k(\mathcal{A}_{\mathcal{X}}, \lbrack \mathcal{Y}_{\mathcal{X}_{l}^t}, \hat{\mathcal{Y}}_{\mathcal{X}_{ul}^t \setminus x}, \hat{y}_{x} \rbrack ) \rbrack, \quad \forall x \in \mathcal{X}_{ul}^t,
\end{aligned}
\end{equation}
where $\hat{\mathcal{Y}}_{\mathcal{X}_{ul}^t \setminus x}$ denotes the r.v. representing the possible labels of ${\mathcal{X}_{ul}^t \setminus x}$ conditioned on $\mathcal{D}_l^t$ . This r.v. is constructed based on Eq. (\ref{eq:met-est}), but we set the data point of interest, $x$, as a variable while fixing other data points, $\mathcal{X}_{ul}^t \setminus x$, as their expected values. Thus, $\tilde{Q}_k (x)$ represents all the possible values of metric $Q_k$ for each data point $x$ when $y_x$ varies.

\textbf{To solve (ii)}, we formulate the acquisition function as the mutual information \citep{Houlsby2011, Gal2017} between $\tilde{Q}_k (x)$ and the BNN weights $\omega$, i.e.,
\begin{equation} \label{eq-IQx}
\mathbb{I}[\tilde{Q}_k (x); \omega \vert x, \mathcal{D}_{l}^t] = \mathbb{H} [\tilde{Q}_k (x) \vert x, \mathcal{D}_{l}^t] - \mathbb{E}_{\omega \sim p(\omega \vert \mathcal{D}_{l}^t)} [\mathbb{H}[\tilde{Q}_k (x) \vert x, \omega]].
\end{equation}

This formulation can be approximated via the BNN posterior distribution $q_{\theta}(\omega \vert \mathcal{D}_{l}^t, \mathcal{S}_l^t)$ using the similar methodology as in \citet{Gal2017}. Detailed formulas are in App. \ref{sec:app-acq-detail}.

\textbf{Finally}, the data point to be sampled is the one with the highest value of $\mathbb{I}[\tilde{Q}_k (x); \omega \vert x, \mathcal{D}_{l}^t]$, i.e.,
\begin{equation} \label{eq_sampmet}
x^*_{t+1} = \text{argmax}_{x \in \mathcal{X}_{ul}^t} \mathbb{I}[\tilde{Q}_k (x); \omega \vert x, \mathcal{D}_{l}^t].
\end{equation}

\subsection{Sampling Methodology for Multiple Metrics} \label{sec-samplelist}
\vspace{-0.2cm}
Given a set of multiple metrics $\{Q_k\}_{k=1}^K$, extending the idea in Sec. \ref{sec:sample-one} to multiple metrics, we can sample the data point with the highest sum of $\mathbb{I}[\tilde{Q}_k (x); \omega \vert x, \mathcal{D}_{l}^t]$, i.e.,
\begin{equation} \label{eq-Iqxmul}
x^*_{t+1} = \text{argmax}_{x \in \mathcal{X}_{ul}^t} \sum\nolimits_{k=1}^K \mathbb{I}[\tilde{Q}_k (x); \omega \vert x, \mathcal{D}_{l}^t],
\end{equation}

\section{Experimental Results}
\vspace{-0.2cm}
\begin{figure}
\begin{center}
\includegraphics[width=0.82\linewidth]{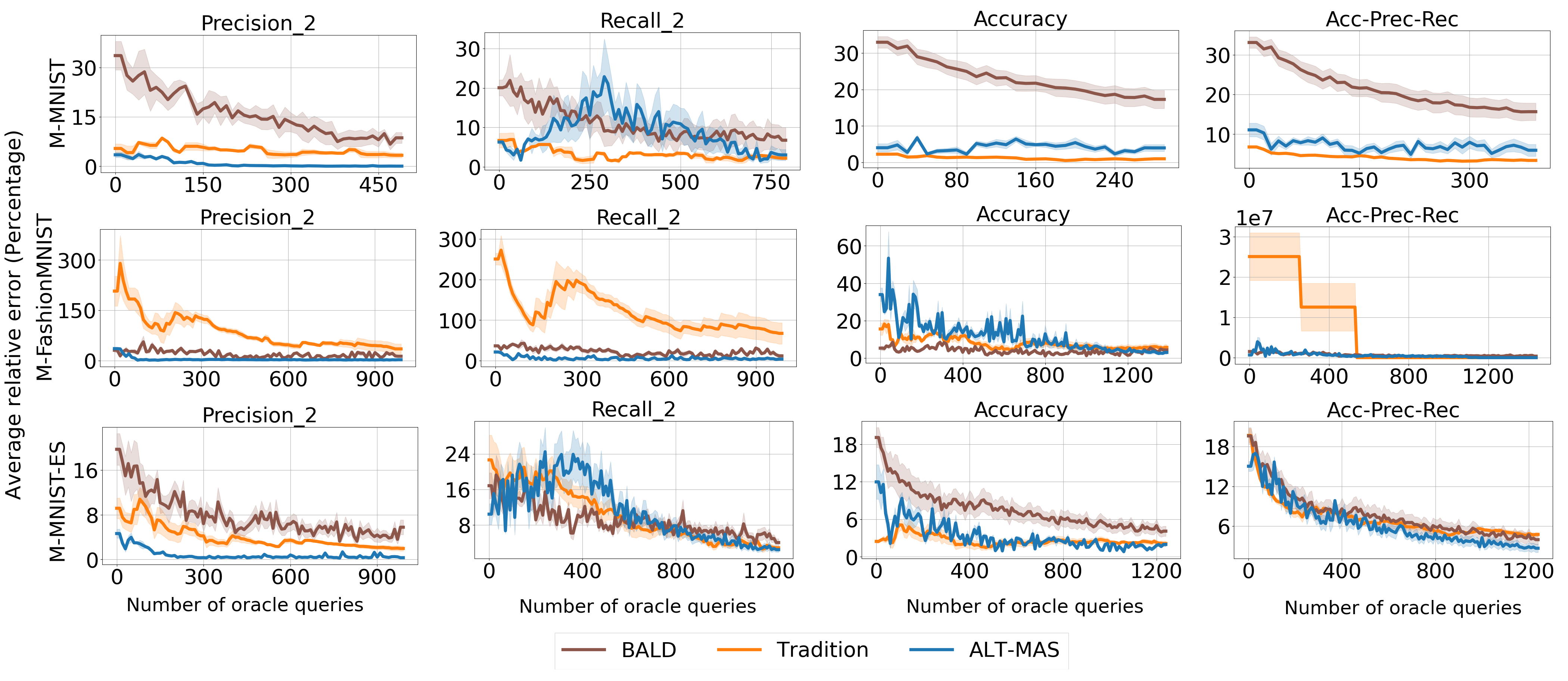}
\vspace{-0.2cm}
\caption{Average relative error of the estimated metrics, for each combination of model-under-test (\textit{M-MNIST}, \textit{M-FashionMNIST}, \& \textit{M-MNIST-ES}) and metric set. Plotting mean and standard error over 3 repetitions. Better methods are those that converge faster to zero (Best seen in color). }
\label{fig:exp-mnist}
\end{center}
\vspace{-0.5cm}
\end{figure}

We evaluate \textbf{ALT-MAS} using various models-under-test and metric sets on MNIST dataset. Our experiments aim to answer the following questions: (Q.1) \textit{Does our active testing framework perform better than traditional machine learning testing approaches?} (Q.2) \textit{Does our active testing framework perform better than existing AL approaches when estimating metric values?} (Q.3) \textit{Does our proposed sampling method work well for different sets of metrics and models-under-test?} and (Q.4) \textit{How accurate the BNNs predictions are?}

To answer Q.1 \& Q.2, we compare \textbf{ALT-MAS} with two baselines: (i) \textbf{Tradition}: the metric values are computed using their mathematical formula with all the labeled data - that is picked randomly from the whole test dataset - up to the current iteration; (ii) \textbf{BALD}: the metric values are estimated using the predicted label of the BNN trained with BALD - the state-of-the-art deep Bayesian AL method \citep{Gal2017}. Note we don't compare with traditional AL methods as it has been shown deep Bayesian AL methods outperform traditional AL methods \citep{Gal2017}. To answer Q.3, for each dataset, we evaluate all methods using various models-under-test and metric sets. In particular, we use three different models-under-test with different levels of accuracy: \textit{M-MNIST} trained on MNIST training set (high accuracy), \textit{M-MNIST-ES} trained on MNIST training set with early stopping (average accuracy), and \textit{M-FashionMNIST} trained on FashionMNIST \citep{Xiao2017} - a completely different dataset (low accuracy). The metric sets consist of either a single or multiple metrics, and the metrics are of different types: per-class metrics and overall metrics. The BNN we use is a simple MLP model with $2$ layers, $256$ neurons/layer, and with dropout applied in each layer. For each combination of model-under-test and metric set, we repeat the experiment $3$ times. Further details on the experiment set up and the answers to Q.4 are in App. \ref{sec:app-expr}.

In Fig. \ref{fig:exp-mnist}, we report the average relative errors of the estimated metrics for all combinations of model-under-test and metric set. First, we can see that \textbf{Tradition} is not consistent; it performs very well on \textit{M-MNIST} but performs badly on \textit{M-FashionMNIST}. Based on our observations, the reason is that \textit{M-MNIST} has very high accuracy (92.8\%), which means its outputs and the true labels are very similar. Thus, randomly picking data points and estimating the metric values via their mathematical formulas results in the metric estimations to be approximately 90-95\%, which are similar to the true values of the metrics. However, for other models-under-test with lower accuracies, this technique will not work. Now let us compare \textbf{ALT-MAS} and \textbf{BALD}. For most cases, \textbf{ALT-MAS} outperforms \textbf{BALD}, especially on per-class metric and \textit{M-MNIST}. This is understandable as i) \textit{M-MNIST} has high accuracy and thus the BNNs trained with \textbf{ALT-MAS} using our data augmentation method have much higher accuracies compared to those with \textbf{BALD}, and, ii) the optimal sampling strategy to achieve accurate estimations for per-class metrics is very different to the optimal sampling strategy to obtain a BNN with high accuracy.


\vspace{-0.2cm}
\section{Conclusion}
\vspace{-0.2cm}
We propose a novel approach to efficiently evaluate the performance of black-box machine learning models. The core idea is to efficiently estimate important metrics of the model being tested based on BNN. We develop a novel data augmentation method for training the BNN to achieve accurate metric estimations. We also devise a novel theoretic information based sampling strategy to sample data points such that the proposed framework can accurately estimate the metrics of interest simultaneously using a minimal number of labeled  data. Experimental results show that our proposed approach works efficiently for estimating multiple metrics and diverse models-under-test.

\bibliography{reference}
\bibliographystyle{iclr2021_conference}

\appendix

\section{Related Work} \label{sec:app-relatedwork}

\paragraph{Classifier Performance Evaluation with Limited Labeled Data} There have been various research works tackling the problem of evaluating classifier performance on an unlabeled test dataset using a limited number of labeled data and a labelling oracle. However, most of these works have focused only on some specific metrics such as accuracy and precision-recall curve. For instance, the approach in \citet{Sawade2010} aims to estimate a risk score, which is a function of the model-under-test output and the ground-truth (akin to metric), by deriving a sampling distribution that minimizes the variance of the risk score. The data points to be sent to the oracle for labelling is selected using this sampling distribution. But, to the best of our knowledge, this approach has only been shown to be tractable when the risk score is the accuracy metric. With the same goal of efficiently estimating classifiers accuracies, \citet{2012Katariya} propose a hashing based stratified sampling method whilst \citet{Kumar2018} derive a stratified sampling strategy that offers optimal variance reduction in the accuracy estimation. To estimate precision-recall or ROC curves of classifiers, \citet{SabharwalS17} suggested PAULA, a precision approximation method using logarithmic annotations, and then ADASTRAT \citep{Sabharwal18}, an adaptive stratified sampling approach. On the other hand, the methods in \citet{2011Druck} and \citet{Welinder2013} can estimate any types of metrics like ours, however, they require information from the classifier internal structure, particularly the softmax outputs. Specifically, \citet{2011Druck} propose to use the softmax output for constructing the stratification function whilst \citet{Welinder2013} use this information to compute the sampling distribution. More recently, \citet{Gopakumar2018} suggested to search for the worst case model performance using limited labeled data, however, we posit that using worst case to assess the goodness of a model-under-test is an overkill because the worst case is often just an outlier. The work in \citet{Schelter2020} learns to validate the model without labeled data by generating a synthetic dataset representative of the deployment data. The restrictive assumption is that it requires domain experts to provide a set of data generators, a task usually infeasible in reality.

\textbf{Trustworthy Machine Learning}. This line of works aims to assess the confidence of deep neural networks in making predictions on test data points \citep{Platt99, Zadrozny2002, Niculescu2005, Guo2017, Gal2016, Lakshminarayanan2017, Hendrycks2017, Jiang2018, Corbiere2019}. For example, \citet{Gal2016} and \citet{Lakshminarayanan2017} estimate the uncertainty of deep neural networks via Bayesian methods so as to return a distribution over the predictions. The work in \citet{Hendrycks2017} uses the softmax probability of the network to detect if a test data point is misclassified or out-of-distribution. A new trust score to understand whether a classifier's prediction for a test data point can be trusted or not is proposed in \cite{Jiang2018}. Most recently, \citet{Corbiere2019} assess the confidence of a model by proposing a new target criterion for model confidence based on the True Class Probability. These methods rely on the training data and/or internal architecture of the network to generate the model confidence score. Our method, in contrast, assumes an already trained and black-box model; our goal is to estimate the performance of the model on various metrics on a new test dataset.


\textbf{Active Learning.} These methods aim to train a machine learning model in a data-efficient way by selecting the most informative data points for which labels should be acquired \citep{Settles2010, Blundell, Gal2017, Kirsch2019}. These AL methods aim to train a model to predict the labels of new data points accurately whilst our method aims to train a model to estimate a specified metric set accurately. We have shown experimentally that for the active testing framework, our proposed method outperforms existing AL methods.

\section{Algorithm Details} \label{sec:app-algo-detail}

\subsection{Background}

\paragraph{Bayesian Neural Networks} Bayesian neural networks (BNNs) are special neural networks that maintain a distribution over its parameters \citep{Mackay1992, Neal1995}. Specifically, given the training data $\mathcal{D}_{tr} = \{x_i, y_i\}_{i=1}^N$, a BNN can provide the posterior distribution $p^*(\omega \vert \mathcal{D}_{tr})$ with $\omega$ being the neural network weights. In practice, performing exact inference to obtain $p^*(\omega \vert \mathcal{D}_{tr})$ is generally intractable, hence we use a variational approximation technique to approximate this posterior. In particular, we employ the MC-dropout method \citep{Gal2016} as it is known to be both scalable and theoretically guaranteed in terms of inferring the true model posterior distribution $p^*(\omega \vert \mathcal{D}_{tr})$. That is, the MC-dropout method is equivalent to performing approximate variational inference to find a distribution in a tractable family that minimizes the Kullback-Leibler divergence to the true model posterior.

\paragraph{Bayesian Active Learning by Disagreement (BALD)} BALD is a sampling methodology in Bayesian active learning, which samples the data point that maximizes the mutual information between the model prediction and the model posterior \citep{Houlsby2011, Gal2017}. To be more specific, let us denote $\mathcal{D}_{l}^t= \{ \mathcal{X}_{l}^t, \mathcal{Y}_{\mathcal{X}_{l}^t} \}$ as the labeled set obtained after iteration $t$ and $\mathcal{B}_{\omega}^t$ as the BNN trained on $\mathcal{D}_{l}^t$ (with parameter $\omega$), then the acquisition function of BALD is defined as,
\begin{equation}
\mathbb{I}[y; \omega \vert x, \mathcal{D}_{l}^t] = \mathbb{H} [y \vert x, \mathcal{D}_{l}^t] - \mathbb{E}_{\omega \sim p(\omega \vert \mathcal{D}_{l}^t)} [\mathbb{H}[y \vert x, \omega]],
\end{equation}
where $y$ is predicted label for the data point $x$. The data point that maximizes BALD is the data point for which the model has many possible predictions, i.e. the posterior draws have disagreement.

\subsection{Data Augmentation Strategy} \label{sec:app-data-aug}

At iteration $t$, given the labeled set $\mathcal{D}_{l}^t$ and the model-under-test outputs $\mathcal{A}_{\mathcal{X}}$, the goal is to train a BNN $\mathcal{B}_{\omega}$ such that the corresponding metric estimations are most accurate. Training BNN using solely the labeled set $\mathcal{D}_{l}^t$ might not result in accurate enough metric estimations. Thus, to improve the metric estimation accuracy, we propose to incorporate the information from the model-under-test outputs $\mathcal{A}_{\mathcal{X}}$ into the BNN training process. In particular, using the labeled set $\mathcal{D}_{l}^t$, we also train a binary classifier $\mathcal{C}_{\eta}$ that aims to predict the data points in the test dataset for which the model-under-test agrees with the ground-truth. Using the predictions by the classifier $\mathcal{C}_{\eta}$, we then construct an \textit{augmented labeled set} $\mathcal{S}_l^t = \lbrace \mathcal{X}_S^t, \mathcal{Y}_{\mathcal{X}_S^t} \rbrace$ where $\mathcal{X}_S^t$ are all the data points in the test dataset $\mathcal{X}$ that $\mathcal{C}_{\eta}$ identifies the model-under-test predictions are accurate, and $\mathcal{Y}_{\mathcal{X}_S^t}$ are the corresponding model-under-test outputs of $\mathcal{X}_S^t$. The BNN is then trained using both the labeled set $\mathcal{D}_{l}^t$ and the augmented labeled set $\mathcal{S}_l^t$.

To train the binary classifier $\mathcal{C}_{\eta}$, we first split the labeled set $\mathcal{D}_{l}^t$ into two parts: training and validation, and then train $\mathcal{C}_{\eta}$ on the training part whilst tuning the softmax probability threshold using the validation part so that $\mathcal{C}_{\eta}$ achieves \textit{the highest precision} on the validation part. This is because we want $\mathcal{C}_{\eta}$ to choose a data point only when it is most certain that the ground-truth and the model-under-test output of that data point is the same. Besides, as the precision of $\mathcal{C}_{\eta}$ is rarely $100\%$, thus, after obtaining the set of data points provided by $\mathcal{C}_{\eta}$, we only take $N_s$ data points from this set with the highest softmax probability. The number $N_s$ is computed by multiplying the precision of $\mathcal{C}_{\eta}$ on the validation part with the cardinality of the original predicted set. For example, if the precision of the classifier $\mathcal{C}_{\eta}$ is $50\%$ on the validation part and the original predicted set consisting of $100$ data points, the final augmented labeled set $\mathcal{S}_l$ only consists of $50$ data points with the highest softmax probability. Finally, the binary classification problem can be imbalanced, particularly when the model-under-test is very accurate or very bad. Hence, when training $\mathcal{C}_{\eta}$, we employ the oversampling technique (for the minority class) to ensure the training data of the binary classification problem to be balanced, i.e. the cardinalities of the majority and minority classes are equal. 

\begin{remark}
\normalfont
With this training methodology, the more accurate the model-under-test, the more accurate the BNN $\mathcal{B}_{\omega}$. In case when the model-under-test is bad, the augmented labeled set $\mathcal{S}_l^t$ does not consist of many elements, thus, the BNN accuracy does not improve much compared to when training solely using the labeled set $\mathcal{D}_{l}^t$. However, in this case, the BNN does not need to have high accuracy in order to accurately estimate the metrics. Specifically, for any data point for which the model-under-test disagrees with the ground truth, the BNN does not need to accurately predict its label. That is, even when the BNN predicts other labels, except the model-under-test output label, the metric estimation is still accurate (more detailed examples in App. \ref{sec:app-BNN}).
\end{remark}

\begin{remark}
\normalfont
For simplicity, we suggest to set the architecture of the binary classifier and BNN to be same. For example, if the BNN is a 2-layer MLP, then the binary classifier is also a 2-layer MLP.
\end{remark}

\begin{remark}
\normalfont
In our implementation, to ensure the augmented labeled set $\mathcal{S}_l^t$ is accurate, we compute $N_s$ to be the square of $C_{\eta}$ precision multiplying with the cardinality of the original predicted set. For example, if the precision of the classifier $\mathcal{C}_{\eta}$ is $50\%$ on the validation part and the original predicted set consisting of $100$ data points, the final augmented labeled set $\mathcal{S}_l$ only consists of $25$ data points with the highest softmax probability.
\end{remark}

\subsection{Details of the Proposed Acquisition Function} \label{sec:app-acq-detail}

Our proposed acquisition function is defined as the mutual information \citep{Houlsby2011, Gal2017} between $\tilde{Q}_k (x)$ and the BNN parameters $\omega$, i.e.,
\begin{equation} \label{eq-IQx-app}
\mathbb{I}[\tilde{Q}_k (x); \omega \vert x, \mathcal{D}_{l}^t] = \mathbb{H} [\tilde{Q}_k (x) \vert x, \mathcal{D}_{l}^t] - \mathbb{E}_{\omega \sim p(\omega \vert \mathcal{D}_{l}^t)} [\mathbb{H}[\tilde{Q}_k (x) \vert x, \omega]].
\end{equation}
This mutual information can be approximated using the MC-dropout variational distribution $q_{\theta}(\omega \vert \mathcal{D}_l^t)$. Next, we show how to approximate each tearm on the right hand side of Eq. (\ref{eq-IQx-app}).

\paragraph{Computing $\pmb{\mathbb{H} [\tilde{Q}_k (x) \vert x, \mathcal{D}_{l}^t]}$} For each data point $x$, $\hat{y}_{x}$ is a discrete random variable with $C$ distinct values, so $\tilde{Q}_k(x)$ is also a discrete random variable with at most $C$ distinct values. Therefore, $\mathbb{H} [\tilde{Q}_k (x) \vert x,  \mathcal{D}_{l}^t]$ can be computed as,
\begin{equation}
\begin{aligned}
\mathbb{H} [\tilde{Q}_k (x) \vert x, \mathcal{D}_{l}^t] &= -\sum\nolimits_{q \in \mathcal{Q}} p(\tilde{Q}_k (x)=q \vert x, \mathcal{D}_{l}^t) \log p(\tilde{Q}_k (x)=q \vert x, \mathcal{D}_{l}^t),
\end{aligned}
\end{equation}
where $\mathcal{Q}$ consists of all the possible values of $\tilde{Q}_k(x)$ when $\hat{y}_x \in \{1, ..., C\}$. By using the union bound, $\mathbb{H} [\tilde{Q}_k (x) \vert x,  \mathcal{D}_{l}^t]$ can then be expressed as,
\begin{equation} \nonumber
\begin{aligned} 
\mathbb{H} [\tilde{Q}_k (x) \vert x, \mathcal{D}_{l}^t] = -\sum\nolimits_{q \in \mathcal{Q}} \Big( \sum\nolimits_{h \in \tilde{Q}_k^{-1}(q)} p(\hat{y}_x = h \vert x, \mathcal{D}_{l}^t) \Big) \log \Big( \sum\nolimits_{h \in \tilde{Q}_k^{-1}(q)} p(\hat{y}_x = h \vert x, \mathcal{D}_{l}^t) \Big),
\end{aligned}
\end{equation}
where $\tilde{Q}_k^{-1}(q)$ is the inverse function that maps the value of $\tilde{Q}_k(x)$ to $\hat{y}_x$. Given $M$ stochastic forward passes $\{\hat{\omega}_j\}$ from the MC-dropout posterior distribution $q_{\theta}(\omega \vert \mathcal{D}_{l}^t)$, $\mathbb{H} [\tilde{Q}_k (x) \vert x, \mathcal{D}_{l}^t]$ can finally be approximated as,
\begin{equation}
\begin{aligned}
\mathbb{H} [\tilde{Q}_k (x) \vert x, \mathcal{D}_{l}^t] \ &\approx -\sum\nolimits_{q \in \mathcal{Q}} \Big( \Big( \sum\nolimits_{h \in \tilde{Q}_k^{-1}(q)} (\sum\nolimits_{j=1}^M p(\hat{y}_x = h \vert x, \hat{w}_j))/M \Big) \\ 
& \quad \quad \quad \quad \times \log \Big( \sum\nolimits_{h \in \tilde{Q}_k^{-1}(q)} (\sum\nolimits_{j=1}^M p(\hat{y}_x = h \vert x, \hat{w}_j))/M \Big) \Big),
\end{aligned}
\end{equation}
where $\tilde{Q}_k(x)$ can be approximated as $\tilde{Q}_k(x) \approx \mathbb{E}_{\omega \sim q_{\theta}(\omega \vert \mathcal{D}_l^t)} \ \lbrack Q_k(\mathcal{A}_{\mathcal{X}}, \lbrack \mathcal{Y}_{\mathcal{X}_{l}^t}, \hat{\mathcal{Y}}_{\mathcal{X}_{ul}^t \setminus x, \omega}, \hat{y}_{x} \rbrack ) \rbrack \approx (\sum\nolimits_{j=1}^M Q_k(\mathcal{Y}_{\mathcal{A}}, \allowbreak [\mathcal{Y}_{l}^t, \hat{\mathcal{Y}}_{\mathcal{X}_{ul}^t \setminus x,\hat{\omega}_j}, \hat{y}_x]))/M$, with $\hat{\mathcal{Y}}_{\mathcal{X}_{ult}^t \setminus x,\hat{\omega}_j}$ denoting the predicted labels for $\mathcal{X}_{ul}^t \setminus x$ given the parameter $\hat{\omega}_j$.

\paragraph{Computing $\pmb{\mathbb{E}_{\omega \sim p(\omega \vert \mathcal{D}_{l}^t)} [\mathbb{H}[\tilde{Q}_k (x) \vert x, \omega]]}$} Similar as in the above paragraph, given $M$ stochastic forward passes $\{\hat{\omega}_j\}_{j=1}^M$ from the MC-dropout variational distribution $q_{\theta}(\omega \vert \mathcal{D}_{l}^t)$, $\mathbb{E}_{\omega \sim p(\omega \vert \mathcal{D}_{l}^t)} [\mathbb{H}[\tilde{Q}_k (x) \vert x, \omega]]$ can be approximated as,
\begin{equation}
\begin{aligned}
\mathbb{E}_{\omega \sim p(\omega \vert \mathcal{D}_{l}^t)} [\mathbb{H}[\tilde{Q}_k (x) \vert x, \omega]] & \approx \dfrac{1}{M} \sum\nolimits_{j=1}^M \mathbb{H}[\tilde{Q}_k (x) \vert x, \hat{\omega}_j] \\
& \approx -\dfrac{1}{M} \sum\nolimits_{j=1}^M \Big( \sum\nolimits_{q \in \mathcal{Q}} p(\tilde{Q}_k (x)=q \vert x, \hat{\omega}_j) \log p(\tilde{Q}_k (x)=q \vert x, \hat{\omega}_j) \Big) \\
& \approx -\dfrac{1}{M} \sum\nolimits_{j=1}^M \Bigg( \sum\nolimits_{q \in \mathcal{Q}} \Big( \sum\nolimits_{h \in \tilde{Q}_k^{-1}(q)} p(\hat{y}_x = h \vert x, \hat{\omega}_j) \Big) \\
& \quad \quad \quad \quad \quad \quad \quad \quad \times \log \Big( \sum\nolimits_{h \in \tilde{Q}_k^{-1}(q)} p(\hat{y}_x = h \vert x, \hat{\omega}_j) \Big) \Bigg),
\end{aligned}
\end{equation}
where $\mathcal{Q}$ consists of all the possible values of $\tilde{Q}_k(x)$ when $\hat{y}_x \in \{1, ..., C\}$, $\tilde{Q}_k^{-1}(q)$ is the inverse function that maps the value of $\tilde{Q}_k(x)$ to $\hat{y}_x$, and $\tilde{Q}_k(x)$ can be approximated as $\tilde{Q}_k(x) \approx \mathbb{E}_{\omega \sim q_{\theta}(\omega \vert \mathcal{D}_l^t)} \ \lbrack Q_k(\mathcal{A}_{\mathcal{X}}, \lbrack \mathcal{Y}_{\mathcal{X}_{l}^t}, \allowbreak \hat{\mathcal{Y}}_{\mathcal{X}_{ul}^t \setminus x, \omega}, \hat{y}_{x} \rbrack ) \rbrack \approx (\sum\nolimits_{j=1}^M Q_k(\mathcal{Y}_{\mathcal{A}}, [\mathcal{Y}_{l}^t, \hat{\mathcal{Y}}_{\mathcal{X}_{ul}^t \setminus x,\hat{\omega}_j}, \hat{y}_x]))/M$, with $\hat{\mathcal{Y}}_{\mathcal{X}_{ult}^t \setminus x,\hat{\omega}_j}$ denoting the predicted labels for $\mathcal{X}_{ul}^t \setminus x$ given the parameter $\hat{\omega}_j$.

\section{Experiment Details} \label{sec:app-expr}

\subsection{MNIST dataset} \label{sec:app-expr-MNIST}

\paragraph{Models-under-test} The three models-under-test we use to evaluate our proposed method are:
\begin{enumerate}[label=(\roman*)]
\item 
\textit{M-MNIST}: A CNN trained on the train MNIST dataset \citep{lecun2010}. We implement the model in tensorflow but use the official network architecture published on keras.com\footnote{\url{https://keras.io/examples/mnist_cnn/}} as there is no official source code on tensorflow.org. The accuracy of this model on the test MNIST dataset is $99.06\%$;
\item
\textit{M-MNIST-ES}: A CNN trained on the train MNIST dataset. The model architecture is same as M-MNIST,
but we use early stopping to decrease the model performance, i.e. we set smaller epochs and higher batch size. The accuracy of this model on the test MNIST dataset is $70.67\%$;
\item
\textit{M-FashionMNIST}: A CNN trained on the train FashionMNIST dataset (i.e. a completely different dataset compared to MNIST) \citep{Xiao2017}. The model is implemented using the official code published on tensorflow.org.\footnote{\url{https://www.tensorflow.org/tutorials/keras/classification}} The accuracy of this model on the test MNIST dataset is $12.39\%$.
\end{enumerate}

\paragraph{Metric sets} The metric sets we use to evaluate our proposed method are:
\begin{enumerate}[label=(\roman*)]
\itemsep0em 
\item
Three sets of metrics consisting of only one metric in each set: precision of class 2, recall of class 2, and overall accuracy. These are common metrics used to evaluate performance of classifiers, and these metrics cover both per-class metrics and overall metrics.
\item
One set of metrics consisting of 21 metrics (accuracy, precision and recall of each class).
\end{enumerate}

\paragraph{The BNN and the binary classifier architecture} The BNN and the binary classifier have the same architecture. The architecture is an MLP with two layers, $256$ neurons/layer, and with dropout applied in each layer. The number of MC-dropout samples are $50$. The initial labeled set $\mathcal{D}_{l}^0$ has $100$ data points randomly sampled from the test dataset. We tuned $3$ hyper-parameters: learning rate, epochs and dropout rate using Bayesian optimization \citep{Snoek2012}. We also reinitialize the BNN after each iteration as in \citep{Gal2017}.

\section{Additional Experiment Results} \label{sec:app-expr-add}

%

\subsection{The Quality of the BNNs} \label{sec:app-BNN}

\begin{figure} [h]
\begin{center}
\includegraphics[width=120mm,height=66mm]{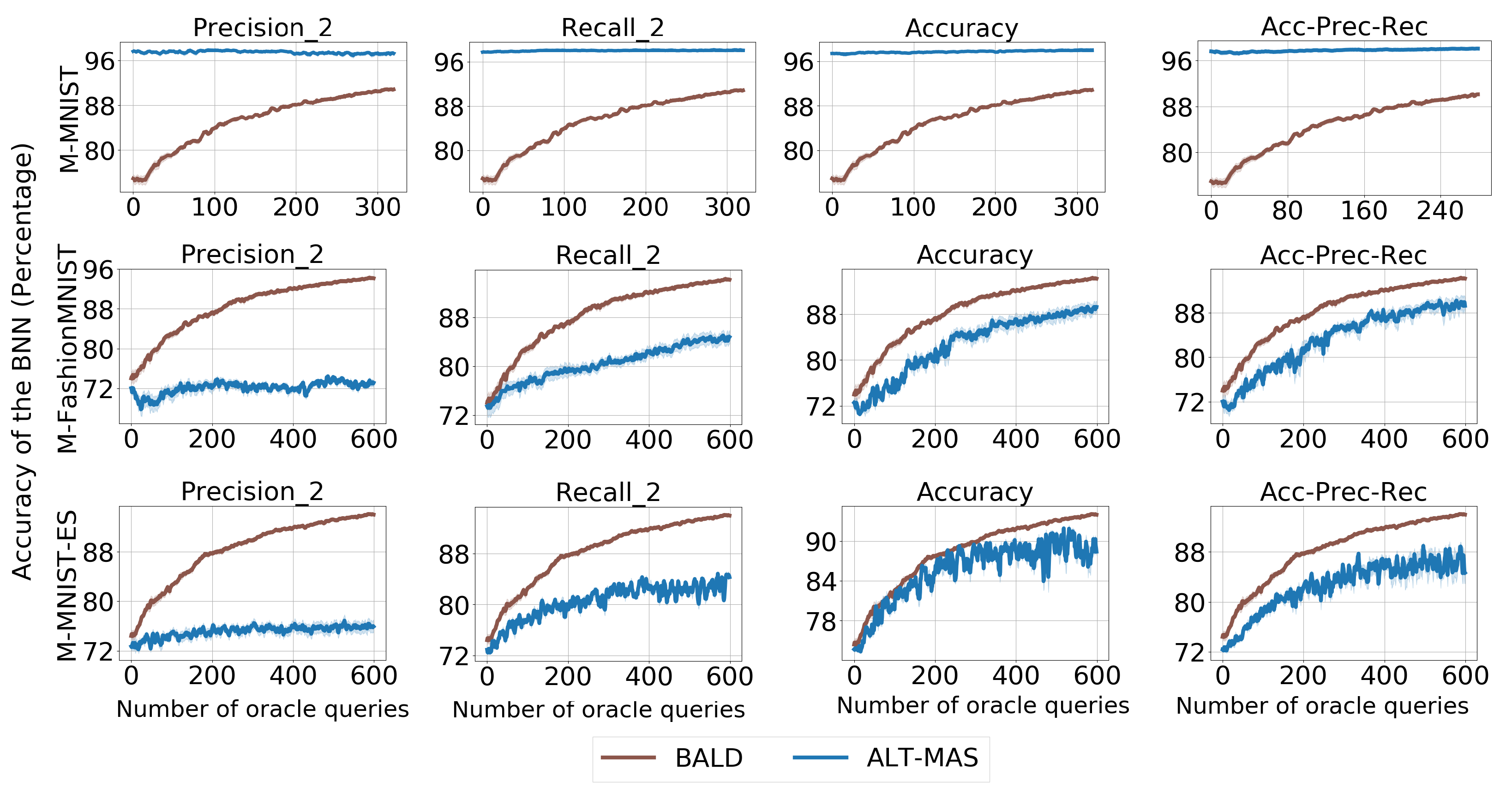}
\caption{The accuracy of the BNN, for each combination of model-under-test (\textit{M-MNIST}, \textit{M-FashionMNIST}, \& \textit{M-MNIST-ES}) and metric set. Plotting mean and standard error over 3 repetitions (Best seen in color). }
\label{results_MNIST_bnn}
\end{center}
\end{figure}

In Fig. \ref{results_MNIST_bnn}, we plot the prediction accuracies of the BNNs in \textbf{BALD} and \textbf{ALT-MAS} for all combinations of models-under-test and metric sets on the MNIST dataset. Below, we evaluate how complex the BNN needs to be for the metric estimation to be accurate and how effective our data augmentation strategy and our proposed sampling methodology are.

Theoretically, we can see that for the BNN to be useful, it does not need to have high classification accuracy; it only needs to accurately predict the data points that contribute to the metric estimation. Being incorrect at other data points does not affect the predicted metric values. See example:

\setlength\tabcolsep{3pt}
\begin{tabular}{ |p{1.0cm}|p{2.5cm}|p{9.5cm}|c| } 
\hline
True label & Model-under-test prediction & Surrogate model prediction \\
\hline
0 & 5 & Any prediction except $5$ (e.g. $1, 2, 3,...$) will give accurate metric estimation \\
\hline 
5 & 5 & Correct prediction matters for some metrics (e.g. accuracy, precision/recall of class 5), but not for other metrics (e.g. precision/recall of other classes) \\ 
\hline
\end{tabular}

In summary, the metric estimation accuracy depends on the model-under-test and the metric set. This is clearly demonstrated by the performance of \textbf{BALD}. To be specific, the BNNs trained with \textbf{BALD} have accuracies ranging from $70-90\%$, but for the models-under-test \textit{M-FashionMNIST} and \textit{M-MNIST-ES} (average \& bad models), the metric estimation accuracies range from $90-100\%$ - which are much higher than the BNNs' accuracies.

For our proposed method \textbf{ALT-MAS}, with the models-under-test \textit{M-FashionMNIST}, \textit{M-MNIST-ES}, the behaviours are similar to those of \textbf{BALD}. That is, the metric estimation accuracies are always higher than the BNNs accuracies, especially for per-class metrics. It is worth noting that, for the per-class metrics, even though the BNNs accuracies by \textbf{ALT-MAS} are much lower than the BNNs by \textbf{BALD}, but the metric estimations by \textbf{ALT-MAS} are much higher than by \textbf{BALD}. This asserts the motivation of our sampling approach, that is, the BNN only needs to accurately predict the data points that contribute to the metric estimation. On the other hand, with the good model-under-test \textit{M-MNIST}, due to our data augmentation training strategy, the BNN accuracies by \textbf{ALT-MAS} are much higher than those of \textbf{BALD}, and thus, the metric estimations by \textbf{ALT-MAS} are also more accurate than those by \textbf{BALD}. 

\end{document}

%% file: math_commands.tex
\usepackage{amsmath,amsfonts,bm}

\def\eqref#1{equation~\ref{#1}}

\def\1{\bm{1}}

\DeclareMathAlphabet{\mathsfit}{\encodingdefault}{\sfdefault}{m}{sl}
\SetMathAlphabet{\mathsfit}{bold}{\encodingdefault}{\sfdefault}{bx}{n}

